\documentclass{article}

\PassOptionsToPackage{numbers, compress}{natbib}
\usepackage[final]{nips_2017}

\usepackage[utf8]{inputenc} 
\usepackage[T1]{fontenc}    
\usepackage{hyperref}       
\usepackage{url}            
\usepackage{booktabs}       
\usepackage{amsfonts}       
\usepackage{nicefrac}       
\usepackage{microtype}      

\usepackage{caption}
\usepackage{subcaption}
\usepackage{graphicx}
\usepackage{algorithmicx}
\usepackage{algorithm}
\usepackage{algpseudocode}
\usepackage[algo2e]{algorithm2e}
\usepackage{color}
\usepackage{times}
\usepackage{url}
\usepackage{amsmath}
\usepackage{amsthm}
\usepackage{amsfonts}
\usepackage{xspace}
\DeclareMathOperator*{\argmax}{arg\,max}

\newcommand{\vt}[1]{
\mathbf{#1}
}

\newcommand{\FreeEnergy}[0]{
  \mathcal{F}
}

\title{Linear-Time Sequence Classification using Restricted Boltzmann Machines}

\author{
  Son N. Tran\\
  The Australian E-Health Research Centre, CSIRO\\
  Brisbane, QLD 2026, Australia\\
  \texttt{son.tran@csiro.au}
   \And
  Srikanth Cherla, Artur d'Avila Garcez, Tillman Weyde\\
  City University London \\
  Northampton Square, London, EC1V 0HB, UK \\
  \texttt{\{srikanth.cherla.1,a.garcez,t.e.weyde\}@city.ac.uk} 
}

\begin{document}

\maketitle

\begin{abstract}
Classification of sequence data is the topic of interest for dynamic Bayesian models and Recurrent Neural Networks (RNNs). While the former can 
explicitly model the temporal dependencies between class variables, the latter have a capability of learning representations. Several attempts have been made to improve performance by combining these two approaches or increasing the processing capability of the hidden units in RNNs. This often results in complex models with a large number of learning parameters. In this paper, a compact model is proposed which offers both representation learning and temporal inference of class variables by rolling Restricted Boltzmann Machines (RBMs) and class variables over time. We address the key issue of intractability in this variant of RBMs by optimising a conditional distribution, instead of a joint distribution. Experiments reported in the paper on melody modelling and optical character recognition show that the proposed model can outperform the state-of-the-art. Also, the experimental results on  optical character recognition, part-of-speech tagging and text chunking demonstrate that our model is comparable to recurrent neural networks with complex memory gates while requiring far fewer parameters.
\end{abstract}
\section{Introduction}
\label{sec:introduction}
Modelling sequences is an important research topic with various applications in audio/music informatics, natural language processing and computer vision. While some work focus on synthesising time-series events \cite{Taylor_2006,Sutskever2007,Sutskever2009}, classification with sequential data also receives much attention \cite{Bengio2003,Nguyen2007,Chen_2015}. For such classification problems, there are two main areas of interest. The first is known as {\it sequence modelling} which is concerned with predicting the next event in time given previous events. The second is {\it sequence labelling}, i.e. how to label a sequence of events, with a wide range of applications such as optical character recognition (OCR) and part-of-speech (POS) tagging.

The classification problems mentioned above are very attractive to the use of dynamic Bayesian models such as Hidden Makov Models (HMMs) \cite{Rabiner_1990} and Conditional Random Fields (CRFs) \cite{Lafferty2001}. An advantage of such models is the ability to learn relationships between sequence labels, which is useful for temporal reasoning. 
Recent research has seen an increasing interest in recurrent neural networks (RNNs) for sequence modelling. Differently from dynamic Bayesian models, RNNs assume that the class labels in a sequence are independent, given the sequence inputs. This makes inference easier, but with the sacrifice of valuable information: the temporal dependencies between sequence labels. A main advantage of RNNs is the ability to learn temporal representations from data using recurrent hidden layers. A major difficulty in this process is learning from long sequences using back-propagation through time \cite{Hochreiter_1997}. This has been addressed mainly by the adoption of complex gates such as in the Long Short Term Memory (LSTM) networks and in networks with Gated Recurrent Units (GRUs) \cite{Cho_2014}.
There are also many attempts to combine the advantages of representation learning with dynamic inference, mostly by integrating a dynamic Bayesian model on top of a deep neural network \cite{Do_2010}. 

In this paper, we propose a novel and compact model based on Restricted Boltzmann Machines (RBMs), which we call Sequence Classification RBMs (SCRBM), to support representation learning and dynamic inference on the classification of sequences. SCRBM is constructed by rolling RBMs with their class nodes over time. Each RBM at time $t$ has a layer of visible units ($X^t$) and a layer of hidden units ($H^t$). Together with class nodes $Y^t$ denoting the labels at time $t$, they form a model representing a distribution: $p(y^{1:T},\vt{x}^{1:T},\vt{h}^{1:T})$ (shorthand for $p(Y^1=y^1,..,Y^T=y^T,X^1=\vt{x}^1,..,X^T = \vt{x}^T,H^1=\vt{h}^1,...,H^T = \vt{h}^T)$). When it comes to inference, there are two questions to answer: first, in order to compute gradients, one needs to infer the hidden states given the state of the input layer and the class labels from the distribution $p(H^{1:T}|\vt{x}^{1:T},y^{1:T})$. This can be done using variational methods, i.e. treating a hidden unit as a mean-field, similarly to \cite{Sutskever2009}. Second,  to predict the state of the class labels given the state of the input layer, in the best case, one can search for the most probable labels from the conditional distribution, i.e. solving $\argmax_{Y^{1:T}} p(Y^{1:T}|\vt{x}^{1:T})$. This may take polynomial time similarly to HMMs and CRFs. Instead, in this paper, we introduce an inference algorithm which only takes linear time. We employ the forward inference in RNNs with the modification that the hidden state at time $t$ is dependent on the previous prediction (class labels at $t-1$). By association, the state of the class labels at any time point is dependent on the prediction of the labels at the previous time point. For learning, we maximise the log-likelihood of the training data. Although it has been shown that learning a local RBM with labels is tractable \cite{Larochelle2008,Srikanth_2017}, learning the entire sequence is not. This is because it is not possible to marginalise out a sequence of hidden units, which is easy in the case of a single discriminative RBM. Also, computation becomes expensive due to the exponential growth of possible assignments to the sequence of labels. To solve this problem, we use the mean-field technique mentioned above to factorise the conditional probability of a sequence into the product of probabilities of local discriminative RBMs, as detailed in the next section. 

Experiments have shown that SCRBM outperforms the state-of-the-art on two benchmark datasets in melody modelling and optical character recognition. Furthermore, we perform extensive evaluations of SCRBM against standard sequence models such as HMMs, CRFs and RNNs. Finally, we also compare SCRBM with RNNs with complex memory gates including GRUs \cite{Cho_2014} and LSTMs \cite{Hochreiter_1997} on the MIT OCR, Penn-Tree bank POS tagging, and Conll2000 Chunking problems. The results show that the performance of the SCRBM model is better than that of standard HMMs and CRFs, and is comparable to that of GRUs and LSTMs while at the same time being considerably more compact with far fewer parameters.

The remainder of the paper is organised as follows. In the next section, we present the theory of {\it sequence classification RBMs}. Then, we discuss the related literature which inspired this work. In section \ref{sec:exp}, we present the empirical evaluations showing the effectiveness of SCRBMs. The last section
concludes the paper and discusses future work.
\section{The SCRBM}
\label{sec:rtdrbm}
\subsection{Model}
The Sequence Classification Restricted Boltzmann Machine (SCRBM) is constructed by rolling RBMs with class labels over time. The model defines a probability distribution: 
\begin{equation}
p(y^{1:T},\vt{x}^{1:T},\vt{h}^{1:T}) = \prod_{t=1}^T p(y^t,\vt{x}^t,\vt{h}^t|\vt{h}^{t-1})
\end{equation}
where $\vt{x}^{1:T}$, $\vt{h}^{1:T}$ are time-series of the visible and hidden states; $y^{1:T}$ is the class-label sequence; $\vt{h}^0$ are the biases of the hidden units.

The main problem of this model, as highlighted in \cite{Sutskever2007}, is that inference is intractable. This, however, can be solved by adding recurrent connections, as done for the Recurrent Temporal Restricted Boltzmann Machine (RTRBM) \cite{Sutskever2009}. In RTRBMs, class labels are not included. In SCRBMs, the local distribution at time $t$: $p(y^t,\vt{x}^t,\vt{h}^t|\vt{h}^{t-1})$ is replaced by:
\begin{equation}
p(y^t,\vt{x}^t,\vt{h}^t|\vt{\hat{h}}^{t-1}) = \frac{\exp(-E_\theta (y^t,\vt{x}^t,\vt{h}^t;\vt{\hat{h}}^{t-1})}{\sum_{y',\vt{x}',\vt{h}'}\exp(-E_\theta(y',\vt{x}',\vt{h}';\vt{\hat{h}}^{t-1})}
\end{equation}
where $\vt{\hat{h}}^{t-1}$ is the expected values of the hidden units at $t-1$:
\begin{equation}
\label{exp}
\vt{\hat{h}}^{t-1} = \mathbb{E}[\vt{H}^{t-1}|\vt{x}^{1:t-1},y^{1:t-1}]
\end{equation}
with local energy function:
\begin{equation}
\begin{aligned}
E_\theta(y^t,\vt{x}^t,\vt{h}^t;\vt{\hat{h}}^{t-1}) = & -[(\vt{x}^t)^\top \vt{W} + \vt{u}_{y^t}^\top  + (\vt{\hat{h}}^{t-1})^\top\vt{W}_{hh}]\vt{h}^t  \\
& - \vt{a}^\top\vt{x}^t - b_{y^t} - \vt{c}^\top\vt{h}^t
\end{aligned}
\end{equation}
which is characterised by the parameters: $\theta = \{\vt{W},\vt{W}_{hh},\vt{U},\vt{a},\vt{b},\vt{c}\}$. Here, $\vt{W}$ is the weight matrix between visible units and hidden units; $\vt{W}_{hh}$  is the recurrent/temporal connection weights of the hidden units; $\vt{U}$ is the weight matrix between the class units (represented by one-hot vectors); $\vt{a}$,$\vt{b}$ and $\vt{c}$ are the biases of the visible units, hidden units and class units respectively; $\vt{u}_{y^t}$ is the ${y^t}^{th}$ column vector of $\vt{U}$. This set of parameters can be reduced by omitting the biases of the visible units because it will be cancelled out in the conditional distribution calculation, as shown in the next section.
\subsection{Inference}
\label{subsec:rbm_disc}
As mentioned earlier, inference of units in the hidden layer given the inputs and labels is easy, as in \cite{Sutskever2009}, where each hidden unit can be treated as a mean-field, as follows: 
\begin{equation}
\label{eq:infer_h_rtdrbm}
\hat{\vt{h}}^t = \sigma(\mathbf{W}^\top \mathbf{x}^t + \mathbf{u}_{y^t} + \vt{W}_{hh}^\top\vt{h}^{t-1} +  \mathbf{c}^{t})
\end{equation}

For classification with SCRBM one would like to search for the most probable assignment of $Y^{1:T}$ given the inputs $x^{1:T}$. Technically, this should be done in a similar way as in HMMs and CRFs. However, in the best case this will take polynomial time. In this paper, we follow the inference mechanism of RNNs which only needs linear time. Note that differently from RNNs, in SCRBM one needs to infer the state of the hidden layer while, at the same time, performing prediction. As  discussed earlier, once the class labels are known, one can infer the state of the hidden layer straightforwardly. We now show that inference of class labels is also easy given the state of the hidden layer. In particular,
from the mean-field values of the hidden layer in the previous time step
we can infer the class labels using the following conditional
distribution:
\begin{equation}
\label{eq:infer_y}
p(y^t|\mathbf{x}^t,\mathbf{\hat{h}}^{t-1}) = \frac{\exp{(-\FreeEnergy(\mathbf{x}^t, y^t,\hat{\mathbf{h}}^{t-1})}}{\sum_{y'}\exp{(-\FreeEnergy(\mathbf{x}^t, y',\hat{\mathbf{h}}^{t-1})}}
\end{equation}

with free energy:
\begin{equation}
\mathcal{F}(\mathbf{x}^t,y,\mathbf{\hat{h}}^t) = -b_{y^t} -\sum_j\log(1+\exp(\mathbf{w}_j^\top \mathbf{x}^t  + u_{yj}+c_j))
\end{equation}

where $\mathbf{w}_j$ is the $j^{th}$ column of the weight matrix
$\vt{W}$ between the hidden units and the units of the visible layer
corresponding to the inputs, and $u_{jy}$ is an element of the weight
matrix $U$ between the hidden units and the units in the visible layer
corresponding to the class labels. Here, the visible biases $\vt{a}$ have been cancelled out which makes the number of parameters equivalent to that of a standard RNN with the same number of hidden units, i.e. $\theta = \{\vt{W},\vt{W}_{hh},\vt{U},\vt{b},\vt{c}\}$.


From \eqref{eq:infer_y} we can predict the value of the class labels. Once
$\vt{y}^{t}$ is known we use it to infer the mean-field values
$\vt{\hat{h}}^{t}$ as in \eqref{eq:infer_h_rtdrbm}. Let us put this in a specific context, starting from $t=1$: the conditional distribution
$p(y^{1}|\vt{x}^{1},\vt{\hat{h}}^{0})$ can be computed exactly
by marginalising out the hidden variable $\vt{h}^1$ while having
$\vt{\hat{h}}^0$ as parameters. In order to calculate from the prediction of the previous step, the values of the current step we do not use the predicted value of $y^{1}$, instead we use the distribution to infer the mean-field value $\vt{\hat{h}}^{1}$. This value is then passed to the next prediction step, and so on. The details of the SCRBM's inference algorithm are given in Algorithm \ref{al1}.

\begin{algorithm}[h]
\KwData{Input: $\vt{x}^{(1:T}$}
 \KwResult{Output: $\vt{y}^{(1:T}$} \hskip .2cm
  \For{$t = 1:T$}{\hskip .15cm
  set $\hat{\vt{y}}^{(t} =  p(\vt{y}^{(t}|\vt{x}^{(t},\vt{\hat{h}}^{(t-1})$ \\
  set $y^{(t} = \argmax_k \hat{y}^{(t}_k $ \\
  set $\hat{\vt{h}}^{(t} = \sigma(\mathbf{W} \mathbf{x}^{(t} + \mathbf{U}\mathbf{\hat{y}}^{(t} +\vt{W}_{hh}\vt{h}^{t-1} + \mathbf{c}^{(t})$ \\ 
  }
\caption{Inference with SCRBM}
\label{al1}
\end{algorithm}
Although this algorithm does not use dynamic programming as in the Viterbi algorithms for HMMs and CRFs, it should still capture the dependencies between class variables over time through the inference of hidden units using the expected values of the class units. 
\subsection{Learning}
\label{learning}
In SCRBMs, we are usually interested in learning the conditional
distribution:
\begin{equation}
\label{}
  \begin{aligned}
     p(y^{1:T}|\mathbf{x}^{1:T}) 
    & = \frac{p(\mathbf{x}^{1:T}, y^{1:T})}
             {\sum_{y'^{1:T}}p(\mathbf{x}^{1:T}, 
              y'^{1:T})} \\
  \end{aligned}
\end{equation}

However, it is difficult to marginalise out all hidden variables $\vt{h}^{1:T}$ to compute exactly this distribution. Moreover, the
complexity of learning our model would increase exponentially with the length of the
sequence, due to the need to sum over all possible classes at every
time step. So, instead of computing the distribution directly we
simplify it by marginalising out the hidden variable at each time $t$
using the expectation of the hidden state at the previous time $t-1$.
Let us consider: 
\begin{equation}
\begin{aligned}
p(\mathbf{x}^{1:T}, y^{1:T}) &= \sum_{\vt{h}^{1:T}}p(y^{1:T}, \mathbf{x}^{1:T},\vt{h}^{1:T})\\
&=\sum_{\vt{h}^{1:T}}\prod_{t=1}^T p(y^{t}, \mathbf{x}^{t},\vt{h}^{t}|\vt{h}^{t-1})\\   
\end{aligned}
\end{equation}

If we first compute the expectation of $\vt{h}^{t-1}$ given the previous input states $\vt{x}^{1:t-1}$ and $y^{1:t-1}$ , which is equivalent to minimising the total energy function of the SCRBM, then we have:
\begin{equation}
q(\mathbf{x}^{1:T}, y^{1:T}) = \prod_{t=1}^T p(y^{t}, \mathbf{x}^{t}|\vt{\hat{h}}^{t-1})
\end{equation}

One can see this as an {e}xpectation step to be followed by an {o}ptimisation step which maximises the log-likelihood of this simplified distribution:
\begin{equation}
  \begin{aligned}
    q(y^{1:T}|\mathbf{x}^{1:T}) & = \frac{\prod_{t=1}^T p(y^{t}, 
              \mathbf{x}^{t}| \hat{\mathbf{h}}^{t-1})}
             {\sum_{y'^{1:T}}\prod_{t=1}^T p(y'^{t}, 
              \mathbf{x}^{t}|\hat{\mathbf{h}}^{t-1})}\\
     &= \frac{\prod_{t=1}^T p(y^{t}, 
              \mathbf{x}^{t}|\hat{\mathbf{h}}^{t-1})}
             {\prod_{t=1}^T \sum_{y'^{t}} p(y'^{t}, 
              \mathbf{x}^{t}|\hat{\mathbf{h}}^{t-1})} \\
             & = \prod_{t=1}^Tp(y^{t}|\mathbf{x}^{t},\hat{\mathbf{h}}^{t-1})
\end{aligned}
\end{equation}

Since $p(y^{t}|\mathbf{x}^{t},\hat{\mathbf{h}}^{t-1})$
is tractable as shown in \eqref{eq:infer_y}, we can compute the above distribution exactly. Now,
one can learn the model by maximising the log-likelihood
function:
\begin{equation} \begin{aligned}
    \ell  = \sum_{y^{1:T},\mathbf{x}^{1:T}}\sum_{t=1}^T \log p(y^{t}|\mathbf{x}^{t},\hat{\mathbf{h}}^{t-1}) \ .
\end{aligned} \end{equation}

Similarly to other time-series connectionist models, such as standard RNNs and RTRBMs, we train the model using back-propagation through time. The update of the model's set of parameters, denoted by $\theta$, is shown below.
\begin{equation}
\begin{aligned}
\nabla \theta&= \sum_{t=1}^T (\frac{\partial_\theta \log p(y^{t}|\mathbf{x}^{t},\hat{\mathbf{h}}^{t-1})}{\partial \theta}  + \frac{\partial_\theta \hat{\mathbf{h}}^{t}}{\partial \theta}\mathcal{O}^t) \\
\end{aligned}
\end{equation}
where $\frac{\partial_\theta \hat{\mathbf{h}}^{t}}{\partial \theta}=\frac{\partial_\theta \sigma(\mathbf{W}^\top \mathbf{x}^t + \mathbf{u}_{y^t} + \vt{W}_{hh}^\top\vt{h}^{t-1} +  \mathbf{c}^{t})}{\partial \theta}$ and $\frac{\partial_\theta \log p(y^{t}|\mathbf{x}^{t},\hat{\mathbf{h}}^{t-1})}{\partial \theta}$ are local derivatives, where $\hat{\mathbf{h}}^{t-1}$ is a value, not a function of $\theta$; and for mathematical convenience,
\begin{equation}
\begin{aligned}
\mathcal{O}^t&= W_{hh}^\top \hat{\mathbf{h}}^{t+1}(1-\hat{\mathbf{h}}^{t+1})\mathcal{O}^{t+1} + \frac{\partial \log p(y^{t+1}|\mathbf{x}^{t+1},\hat{\mathbf{h}}^{t})}{\partial \hat{\mathbf{h}}^{t}}
\end{aligned} 
\end{equation}
 
\section{Related work}
Although SCRBM is a sequence learning connectionist network having the same set of parameters as a recurrent neural network, they are much different in terms of structure, learning and inference, as showed above. 
Modelling sequence data with RBMs has been studied previously \cite{Sutskever2007,Sutskever2009}. However, as in generative models, they are not easy to apply to classification tasks. The key problem is that the exact gradient can not be computed analytically which requires the use of approximation algorithms. In contrast, SCRBM is a discriminative model whose log-likelihood is tractable. Besides such tractability, another motivation for having a discriminatively learned variant of RTRBMs is the desire for better classification performance. It has been shown in \cite{Ng2001} that with
sufficient training examples, discriminative learning tends to do better on the task it is optimized
for than its generative counterpart on the same model. This has been confirmed by the
findings in \cite{Larochelle2008}. In more recent work on the Recurrent Temporal Discriminative Restricted Boltzmann Machines (RTDRBM) \cite{Cherla_2015}, a generalised version of the RTRBM has been tailored with discriminative inference and learning specifically for melody modelling. Here, we make that work, otherwise focused on such application, applicable to other sequence learning tasks, especially sequence labelling as seen in the following section. Finally, RTDRBMs require more parameters than SCRBMs because it generalises RTRBMs by including connections from previous hidden layers to the current class labels.
Another, even more recent related model is the Dynamic Boltzmann Machine \cite{Dasgupta_2017} which supports online learning for sequence prediction, mainly applied to regression. 
\section{Experiments}
\label{sec:exp}
\subsection{Music Modelling}
In the task of music prediction addressed here, one seeks to predict the next pitch in a sequence which is a time series representation of the music. We use the benchmark from \cite{Cherla_2015} which consists of 8 folk and chorale melody datasets. For the evaluation we use cross-entropy to measure the mean
divergence between the entropy calculated from the predicted
distribution and the correct pitch. Given a test data
$\mathcal{D}_{test}$, the cross-entropy can be computed over all the
events belonging to different sequences as follows:
\begin{equation}
  \label{eq:cross_entropy}
  H_c(p_{mod}, \mathcal{D}_{test}) = \frac{-\sum_{s \in
      \mathcal{D}_{test}} \sum_{t=1}^{T_s} \log_2
    p_{mod}(s^{t}|s^{1:t-1})} {\sum_{s \in
      \mathcal{D}_{test}} T_s}
\end{equation}
where $p_{mod}$ is the probability assigned by the model to the pitch
of the next musical event $s^{t}$ in the melody, $s \in
\mathcal{D}_{test}$, given its preceding pitches, and $T_s$ is the
length of $s$. To apply SCRBM to this task we follow the standard encoding in \cite{Bengio2003} where $y^t=s^t$ and $x^t = s^{t-1}$.

We compare the performance of SCRBMs with the
 state-of-the-art approaches on this dataset which include RTRBMs and five
 other connectionist models: standard feedforward neural networks (FNN) \cite{Bridle1990}, Restricted Boltzmann Machines (RBM) \cite{Cherla_2015}, Discriminative RBMs (DRBM) \cite{Larochelle2008}, a bounded order $n$-gram model {\it C2I}, and an unbounded order $n$-gram model {\it C*I}, as well as standard recurrent neural networks (RNNs) and the generalised RTRBM for melody prediction (RTD-RBM) \cite{Cherla_2015}.

For fairness of comparison, we partition the data exactly as done in previous work
\cite{Pearce2004,Cherla2013}. In particular, a resampling method is employed for evaluation of the $8$ datasets \cite{Dietterich1998}. For model selection, we use a small part of the data
($5\%$) in each case as validation set. In order to determine the best hyperparameters for a model a grid search is carried out. Such hyperparameters include the learning rate, regularization parameters and 
number of hidden units.  

\begin{table}
  \centering
  {\scriptsize
  \begin{tabular}{lcc}
    \hline \\
    Model & Context & Cross Entropy (bits)\\
    \hline \\
    \textbf{SCRBM}                      & n/a & $\mathbf{2.712}$ \\
    RTDRBM                         & n/a & $2.741$ \\ 
    RTRBM                                & n/a & $2.764$ \\ 
    RNN$_{n_{h}=100, h_{act}=TanSigU}$ & n/a & $2.787$ \\ 
    RBM$_{n_{h}=100, h_{act}=LogSigU}$ & $8$ & $2.799$ \\ 
    DRBM$_{n_{h}=25, h_{act}=LogSigU}$ & $5$ & $2.819$ \\ 
    FNN$_{n_{h}=200, h_{act}=ReLU}$ & $7$ & $2.819$ \\
    $n$-gram (u) & n/a & $2.878$ \\ 
    $n$-gram (b) & $2$ & $2.957$ \\ 
    \hline \\
  \end{tabular}
  }
  \caption{Test-set performance comparison on pitch prediction time series; $n_{h}$ denotes the number of hidden neurons and $h_{act}$ the activation function used, either standard sigmoid, hyperbolic tangent or rectified linear functions. The context is the size of the window in the time-series for the models that require this to be predefined.}
  \label{tab:results_melody}
\end{table}

Table \ref{tab:results_melody} shows the best predictive performance of each model averaged across all 8 datasets. It indicates a consistent improvement in best-case performance from the n-gram models, the non-recurrent neural networks, and then the recurrent neural network models, with the SCRBM outperforming all of the others. The hyperparameter with the most influence in the result was the size of the hidden layer. It was found both in the case of the RTRBM and the SCRBM that a hidden layer size of 100 units resulted in the best predictive performance.

\subsection{Optical Character Recognition}
\label{subsubsec:dataset_ocr}
The MIT OCR dataset\footnote{http://www.seas.upenn.edu/\textasciitilde
  taskar/ocr/} is a widely used benchmark for evaluating
  sequence labelling algorithms \cite{Taskar2004}. We use two popular partitions from \cite{Nguyen2007} and \cite{Do_2010,Chen_2015}. In the former, called here "ms" for model selection, the data is partitioned into 10 groups, each consisting of a training, validation and test set. We select models based on performance on the validation sets and report their average accuracy on the test sets. In the latter, here called "cv" for cross-validation, the data is divided into 10 folds in the usual way but without model selection. 

Each model is expected to predict the correct label
corresponding to the image of a character as it is drawn. All the models are evaluated using the averaged loss function per sequence $E(y,
y^*)$, given by:
\begin{equation}
  E(y, y^*) = \frac{1}{N} \sum_{i=1}^N \left[ \frac{1}{L_i}
    \sum_{j=1}^{L_i} \mathcal{I}((y_i)_j \neq (y^*_i)_j)\right]
\end{equation}
where $y$ and $y^*$ are the predicted and the true
sequence respectively, $N$ is the total number of test examples, $L_i$
is the length of the $i^{th}$ sequence, and $\mathcal{I}$ is the $0-1$
loss function.
\subsubsection{Baselines}
\label{subsubsec:baseline_ocr}

We compare the performance of SCRBM on the sequence labelling task with the baseline models: SVM-$multiclass$, SVM-$struct$, Max-margin Markov network (M3N), Averaged Perceptron, Search-based structured prediction model (SEARN), CRF, HMM, LogitBoost, TreeCRF, RTDRBM\footnote{This version of RTDRBM is modified by using our inference algorithm to work with sequence labelling task.}; and the state-of-the-arts: 
\begin{itemize}
\item 
Structured learning ensemble (SLE) \cite{Nguyen2007}: An optimised ensemble of 7 effective models: SVM-$multiclass$, SVM-$struct$, M3N , Perceptron, SEARN, CRF, HMM.  
\item 
Neural CRF \cite{Do_2010}: A combination of CRF and deep networks. 
\item
  Gradient boosting CRF (GBCRF) \cite{Chen_2015}: CRF trained by a novel gradient boosting algorithm.
\end{itemize}

\subsubsection{Results}
\label{subsubsec:results_ocr}
For the model selection ("ms") partition, in order to determine the best model for the task, a grid search was carried out. For fair comparison, we use the best results of the baselines  reported in \cite{Nguyen2007} on the same dataset. For SCRBM, RTDRBM and RNN  the hyper-parameters include: the learning rate and the number of hidden unit. 
We also use early stopping in which the performance of the model on a validation set was determined after every $10$ epochs.  If the performance in one of these checks happened to be worse than the previous best one for $5$ consecutive checks the training was stopped. In the sequence learning, stochastic gradient descent  with backpropagation through time was used.

For the cross-validation ("cv") partition, since model selection is not applicable, we fix our SCRBM with $1000$ hidden units and use adaptive learning rate (Adam \cite{Kingma_2014}) started from $0.001$ to get rid of the search for hyper-parameters.

Table \ref{tab:ocr_results} shows the comparative performance of SCRBM.  It is evident that SCRBM clearly outperforms the other models. It is worth noting that the performance of our model is much better than  SLE, Neural-CRF, and GBCRF.
\begin{table}
  \centering
  \begin{tabular}[h]{l|c|c|c|}
    \hline
    \textbf{Model} & ms  & cv  \\
    \hline
    $\mathbf{SCRBM}$  & $\mathbf{14.81}$  & $\mathbf{04.03}$ \\
    RTDRBM & $15.46$ & -\\
    Neural CRF$^{CML}$ & -        & $04.44$\\
    Neural CRF$^{LM}$  & -        & $04.56$\\    
    SLE                & $20.58$  & -\\
    GBCRF              & -        & $04.64$\\
    TreeCRF            & -        & $06.99$\\
    LogitBoost         & -        & $09.67$\\
    RNN                & $22.92$  & $13.30$\\
    M3N                & $25.08$  & $13.46$\\ 
    Perceptron         & $26.40$  & -\\
    SEARN              & $27.02$  & -\\
    SVM$_{multiclass}$ & $28.54$  & -\\
    SVM$_{struct}$     & $21.16$  & -\\
    HMM                & $23.70$  & -\\     
    CRF                & $32.30$  & $14.20$\\
    \hline
  \end{tabular}
  \vskip -.1cm
  \caption{Test errors ($\%$) of different models in the evaluation.}
  \label{tab:ocr_results}
\vskip -.2cm
\end{table}

\subsection{SCRBM v.s LSTM \& GRU}
\begin{figure*}[h]
\centering
\begin{tabular}{ccc}
 \includegraphics[scale=0.22]{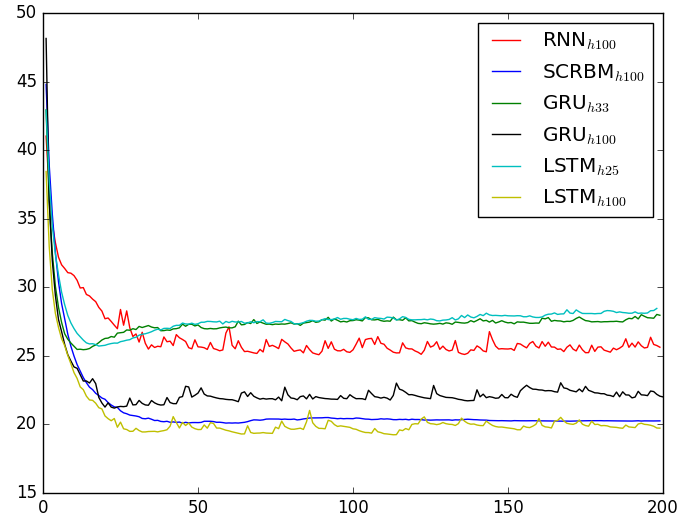} & \includegraphics[scale=0.22]{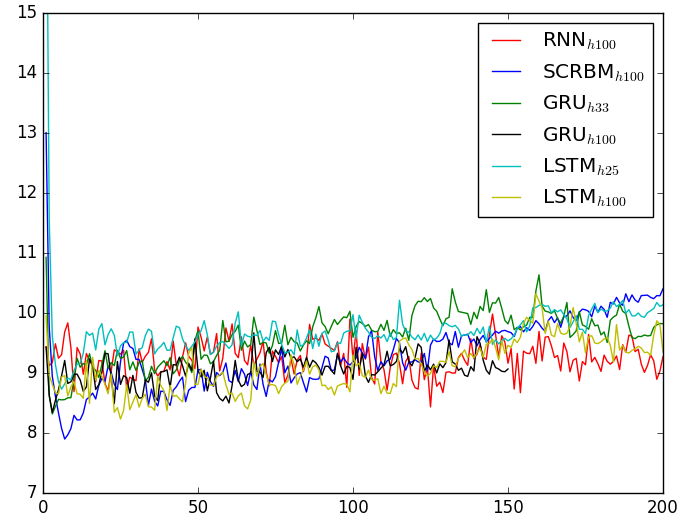} & \includegraphics[scale=0.22]{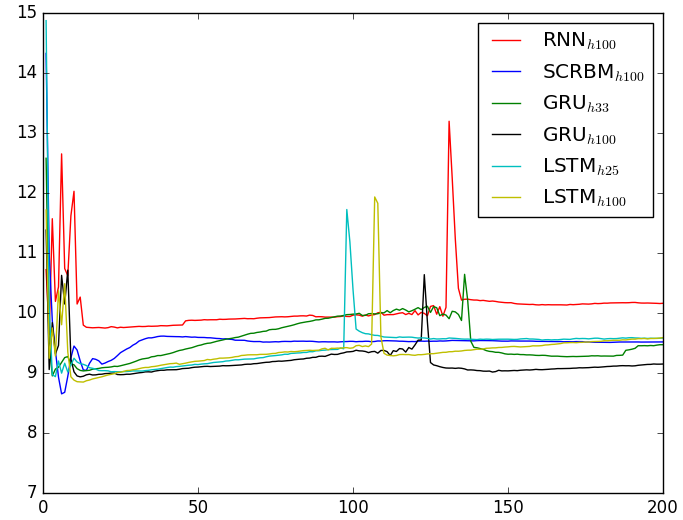} \\
(a.1) OCR validation. & (b.1) POS validation. & (c.1) Chunking validation. \\
\includegraphics[scale=0.22]{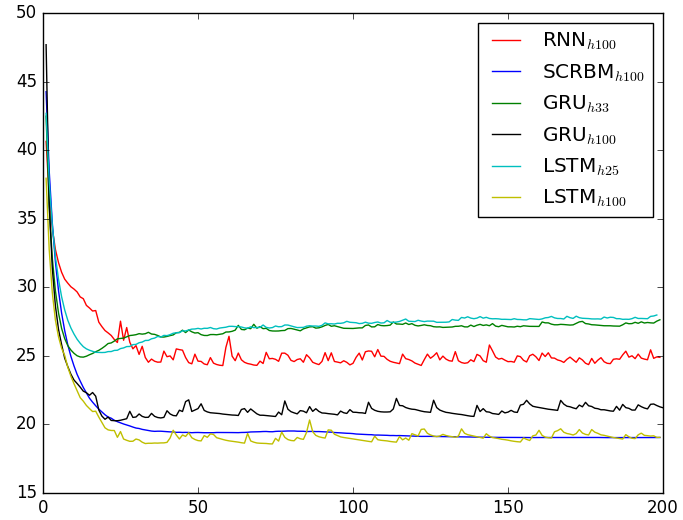} & \includegraphics[scale=0.22]{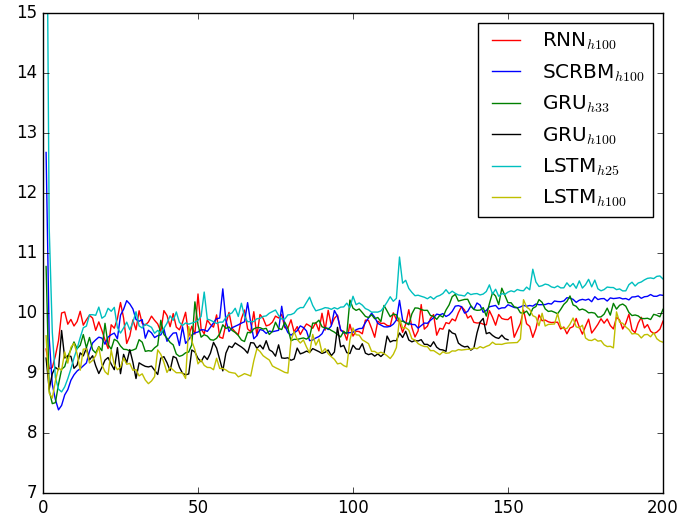} & \includegraphics[scale=0.22]{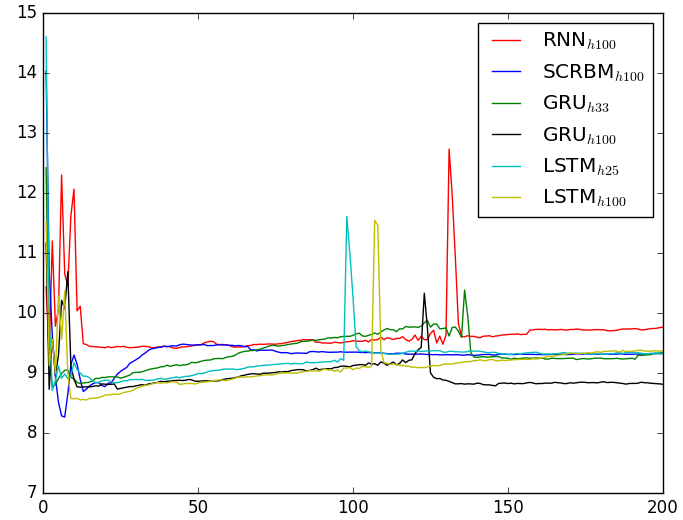} \\
(a.2) OCR test. & (b.2) POS test. & (c.2) Chunking test. 
\end{tabular}
\vskip -.15cm
\caption{Prediction errors. Averaged negative log-likelihood. x-axis: number of epoch.y axis: prediction errors.}
\label{fig:err}
\end{figure*}
\begin{figure*}[h]
\vskip -.2cm
\centering
\begin{tabular}{ccc}
 \includegraphics[scale=0.22]{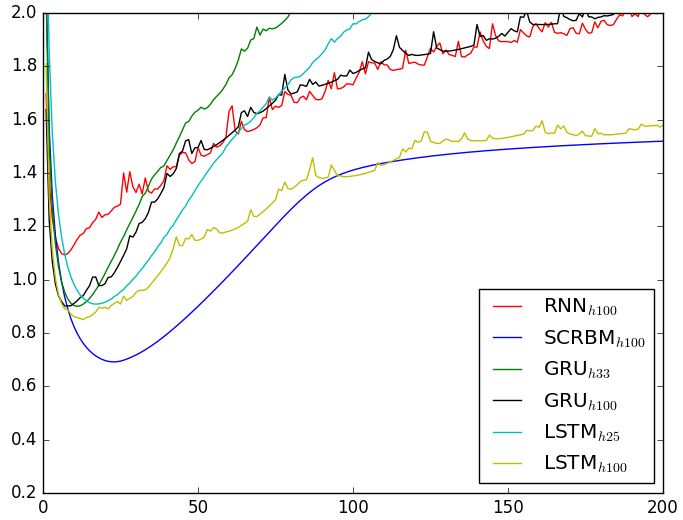} & \includegraphics[scale=0.22]{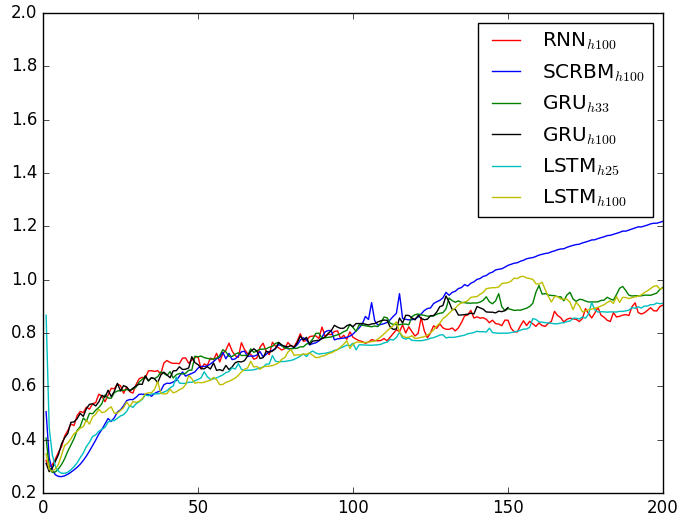} & \includegraphics[scale=0.22]{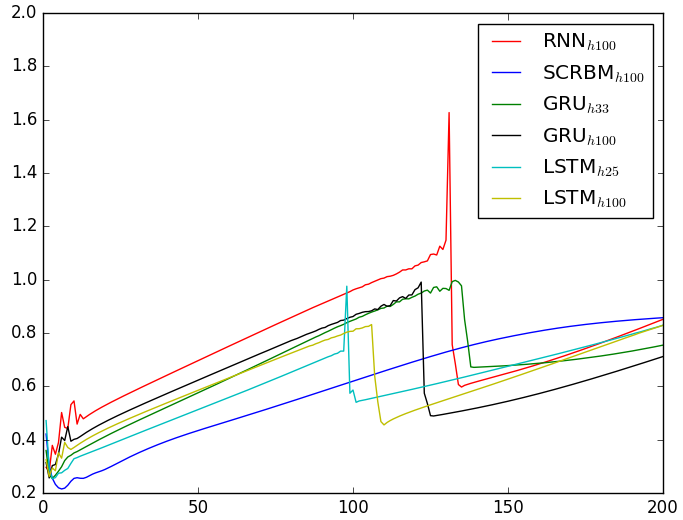} \\
(a.1) OCR validation. & (b.1) POS validation. & (c.1) Chunking validation. \\
\includegraphics[scale=0.22]{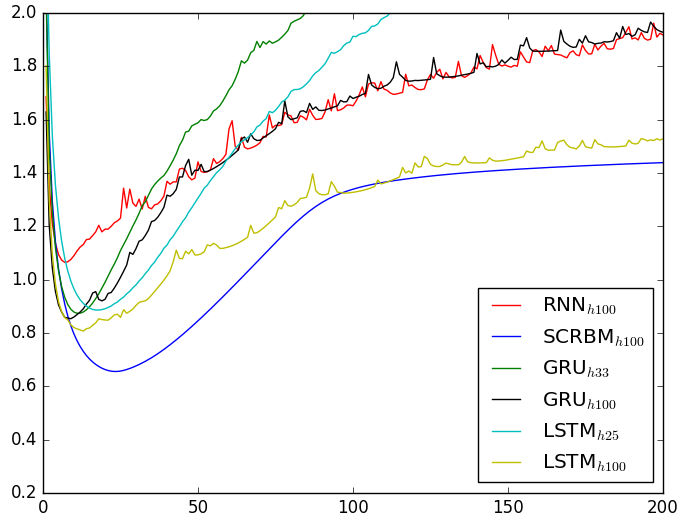} & \includegraphics[scale=0.22]{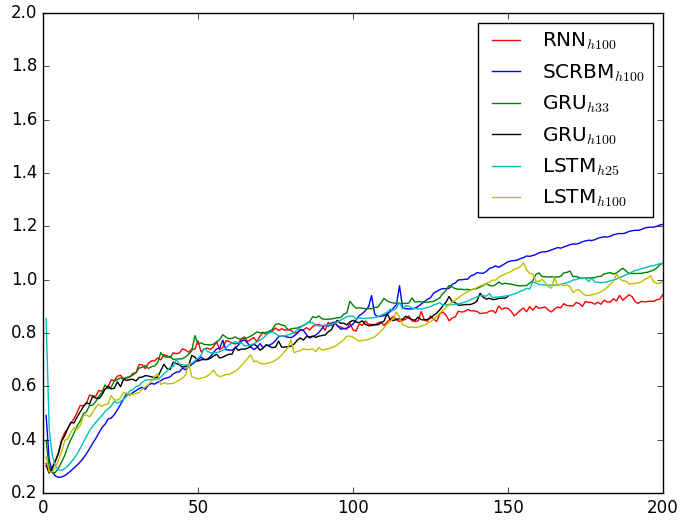} & \includegraphics[scale=0.22]{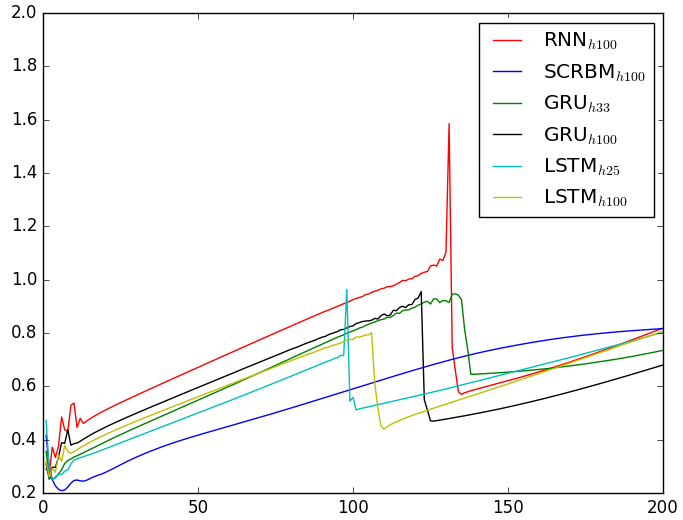} \\
(a.2) OCR test. & (b.2) POS test. & (c.2) Chunking test
\end{tabular}
\vskip -.15cm
\caption{Averaged negative log-likelihood. x-axis: number of epoch.y axis: average negative log-likelihood.}
\label{fig:nllh}
\vskip -.2cm
\end{figure*}
For completeness, we now compare our model with different recurrent
neural networks, especially with ones having complex memory gates in hidden units such
as {\it gated recurrent units} (GRU) and {\it long short term memory} (LSTM).  We
use a traditional RNN with tanh hidden units (denoted as RNN), HMMs, and CRFs as baselines. 

It is worth noting that for a visible layer with $M$ units and a hidden layer with $N$ units, the number of weights of RNN and RTDRM are the same while  GRU and LSTM have $2N(M+N+1)$and $3N(M+N+1)$  more weights than the RNN and SCRBM, respectively. In general, if the number of labels is small then SCRBM is $\approx 3$ times smaller than GRU and $\approx 4$ times smaller than LSTM.
\begin{table}[ht]
\centering
{\scriptsize
\begin{tabular}{|c|c|c|c|c|c|c|c|}
\hline
& OCR & POS$_{500}$ & POS$_{1000}$ & POS$_{2000}$ & Chunking\\
\hline
\hline
HMM    & 23.700 & 23.460 & 19.950 & 17.960 & -\\
CRF    & 32.300 & 16.530 & 12.510 & 09.840 & -\\
RNN    & 22.921 & 12.220 & 10.480 & 09.207 & 09.303\\
GRU    & 15.853 & 11.663 & 09.892 & 09.011 & 08.759\\
LSTM   & 13.329 & 12.012 & 10.168 & 09.273 & 08.561\\
SCRBM & 14.808 & 12.119 & 10.061 & 09.178 & 08.286\\
\hline
\end{tabular}
}
\caption{Test errors of HMMs, CRFs, RNNs: recurrent neural network with $tanh$ hidden units,GRU: gated recurrent units, LSTM: long short term memory. Model sizes and other hyperparameters are opimised in a grid search}
\label{tab:pos}
\end{table}

We test both GRU and LSTM on the OCR dataset with "ms" partition  above. Furthermore, we also perform experiments with POS tagging dataset from Penn Treebank \footnote{http://www.cis.upenn.edu/ treebank}. The data is partitioned into different training sets of $500$, $1000$, and $2000$ samples. The models are selected by using a held-out $10\%$ samples of training sets, which are then evaluated in a test set of $\sim 1600$ sentences. The challenge of this data is that the lexical features are very large, around $450,000$. The final dataset is used for evaluation here is Conll200 chunking\footnote{https://www.clips.uantwerpen.be/conll2000/chunking/}. Since the original data does not provide a validation set, we take $2.5\%$ of the training samples for training, the rest is for validation. The original test set is used for testing. The results are shown in Table \ref{tab:pos}. We can see that  our SCRBM outperforms  the standard models: RNNs, CRFs, and HMMs. In comparison to RNNs with complex gates, SCRBM has comparable performance as GRU and LSTM. In particular, SCRBM is better than GRU in the OCR dataset and better than LSTM in the POS dataset. It is more effective than both GRU and LSTM in the Conll2000 chunking. We should note that even though the LSTM seems to beat our SCRBM in OCR dataset it needs $2000$ hidden units for that best result while SCRBM only requires $100$ hidden units.

Finally we compare the effectiveness of SCRBM, RNN, GRU and LSTM having the same number of parameters. We use $100$ hidden units for both SCRBM and RNN, $33$ hidden units for GRU and $25$ hidden units for LSTM. We even include GRU and LSTM with $100$ hidden units in the test to give a more comprehensive  comparison. We evaluate the models on the OCR, POS tagging with $2000$ training sentences, and Conll 2000 chunking. The evaluation metrics are prediction error rate and the average negative log-likelihood on validation and test sets. All models are trained using Adam with learning rate $0.001$ which, as we found, is generally good for all models on three datasets. Both Figure \ref{fig:err} and Figure \ref{fig:nllh} show that SCRBM generalises better than RNNs, GRU and LSTM which have the same size or same number of hidden units. For prediction errors, Figures \ref{fig:err}b.1, \ref{fig:err}b.2, \ref{fig:err}c.1, \ref{fig:err}c.2 demonstrate that in POS and Chunking datasets  all models converge very quickly after few epochs with the SCRBM reaches the lowest error. We can see from Figure \ref{fig:err}a.1 In OCR dataset, SCRBM performs similarly as  LSTM with the same number of 100 hidden units. In the case of the average negative log-likelihood, its value is inversely proportional to the cost function we want to maximise in the training of all models. Therefore, as we can see in Figures \ref{fig:nllh}a.1, \ref{fig:nllh}a.2, \ref{fig:nllh}b.1, \ref{fig:nllh}b.2, \ref{fig:nllh}c.1, \ref{fig:nllh}c.2 that the SCRBM generalises
better than other models as it achieves the lowest negative loglikelihoods
on the validation and test sets. This, along with the Table \ref{tab:pos}, explains why in some cases GRUs and LSTMs may be better but only when they have more hidden units.  Also, the Figures \ref{fig:err}, \ref{fig:nllh} make the comparison more complete by answering a question on our claim about compactness, e.g what if GRU and LSTM are better with the same number of parameters as SCRBM.
\section{Conclusions}
\label{sec:conclusions}
In this paper, we propose a novel model for classification  with
sequences by rolling restricted Boltzmann
machines with class layer over time. We name it as {\it Sequence classification RBMs}. The key characteristic of this SCRBM is that it can support both representation learning and temporal reasoning. Also, the proposed model is very compact, its number of parameters is equivalent to a basic recurrent neural network having the same number of hidden units.  Inference with SCRBM is done by
performing prediction of the class and computing mean-field values of
hidden layer at each time $t$, consecutively. We train SCRBM by computing the expectation of hidden units at time $t-1$ for each element of the conditional distribution which makes learning tractable.  In the experiments, we evaluate the
effectiveness of SCRBM in sequence modelling with a benchmar of 8 melody datasets
and in sequence labelling with two different partitions of OCR dataset. In both cases, SCRBM outperforms the state-of-the-arts. We also show the
simplicity advantage of SCRBM over GRU and LSTM, two types of
recurrent neural networks with complex memory gates in hidden units.

\small
\bibliographystyle{plain}
\bibliography{biblio}

\end{document}